\documentclass[letterpaper, 10 pt, conference]{ieeeconf}  

\usepackage[colorlinks=true,linkcolor=black,anchorcolor=black,citecolor=black,filecolor=black,menucolor=black,runcolor=black,urlcolor=black]{hyperref}

\IEEEoverridecommandlockouts                              

\makeatletter
\let\NAT@parse\undefined
\makeatother
\usepackage{apacite}

\usepackage{graphicx}
\usepackage{caption}
\usepackage{subcaption}
\usepackage{mathtools}
\usepackage{hyperref}

\usepackage{amsmath} 
\usepackage{amssymb}  

\usepackage{amsfonts}
\usepackage{booktabs}
\usepackage{array}
\newcolumntype{P}[1]{>{\centering\arraybackslash}p{#1}}

\usepackage{stackengine}

\def\invcross{%
  \stackon[0.5ex]{\rule{0.4pt}{1.5ex}}{\rule{.75ex}{0.4pt}}}

\title{\LARGE \bf
Real-Time Reinforcement Learning for Vision-Based Robotics\\ Utilizing Local and Remote Computers
}

\author{Yan Wang$^{*\dag}$, Gautham Vasan$^{*\dag}$, A. Rupam Mahmood$^{\dag\invcross}$ 
\thanks{*Equal Contribution. 
\newline $^{\dag}$Department of Computing Science, University of Alberta, Edmonton AB., Canada, T6G 2E8
\newline $^{\invcross}$ CIFAR AI Chair, Alberta Machine Intelligence Institute (Amii)
\newline Email: {\tt$\{$ \small yan28, vasan, armahmood$\}$@ualberta.ca}
\newline Video: \url{https://youtu.be/7iZKryi1xSY}
}
}

\begin{document}
\maketitle
\thispagestyle{empty}
\pagestyle{plain}

\begin{abstract}

Real-time learning is crucial for robotic agents adapting to ever-changing, non-stationary environments. 
A common setup for a robotic agent is to have two different computers simultaneously: a resource-limited local computer tethered to the robot and a powerful remote computer connected wirelessly.
Given such a setup, it is unclear to what extent the performance of a learning system can be affected by resource limitations and how to efficiently use the wirelessly connected powerful computer to compensate for any performance loss.
In this paper, we implement a real-time learning system called the \emph{Re}mote-\emph{Lo}cal \emph{D}istributed (ReLoD) system to distribute computations of two deep reinforcement learning (RL) algorithms, Soft Actor-Critic (SAC) and Proximal Policy Optimization (PPO), between a local and a remote computer.
The performance of the system is evaluated on two vision-based control tasks developed using a robotic arm and a mobile robot.
Our results show that SAC's performance degrades heavily on a resource-limited local computer.
Strikingly, when all computations of the learning system are deployed on a remote workstation, SAC fails to compensate for the performance loss, indicating that, without careful consideration, using a powerful remote computer may not result in performance improvement.
However, a carefully chosen distribution of computations of SAC consistently and substantially improves its performance on both tasks.
On the other hand, the performance of PPO remains largely unaffected by the distribution of computations.
In addition, when all computations happen solely on a powerful tethered computer, the performance of our system remains on par with an existing system that is well-tuned for using a single machine.
ReLoD is the only publicly available system for real-time RL that applies to multiple robots for vision-based tasks. The source code can be found at \url{https://
github.com/rlai-lab/relod}


\end{abstract}

\section{INTRODUCTION}

Building robotic agents capable of adapting to their environments based on environmental interactions is one of the long-standing goals of embodied artificial intelligence. 
Such a capability entails learning on the fly as the agent interacts with the physical world, also known as \textit{real-time learning}. 
When learning in real-time, the real world does not pause while the agent computes actions or makes learning updates (Mahmood et al.\ 2018a, Ramstedt \& Pal 2019).
Moreover, the agent obtains sensorimotor information from various onboard devices and executes action commands at a specific frequency. 
Given these constraints, a real-time learning agent must compute an action within a chosen action cycle time and perform learning updates without disrupting the periodic execution of actions (Yuan \& Mahmood 2022).

\begin{figure}[t]
\centering
\includegraphics[width=0.375\textwidth]{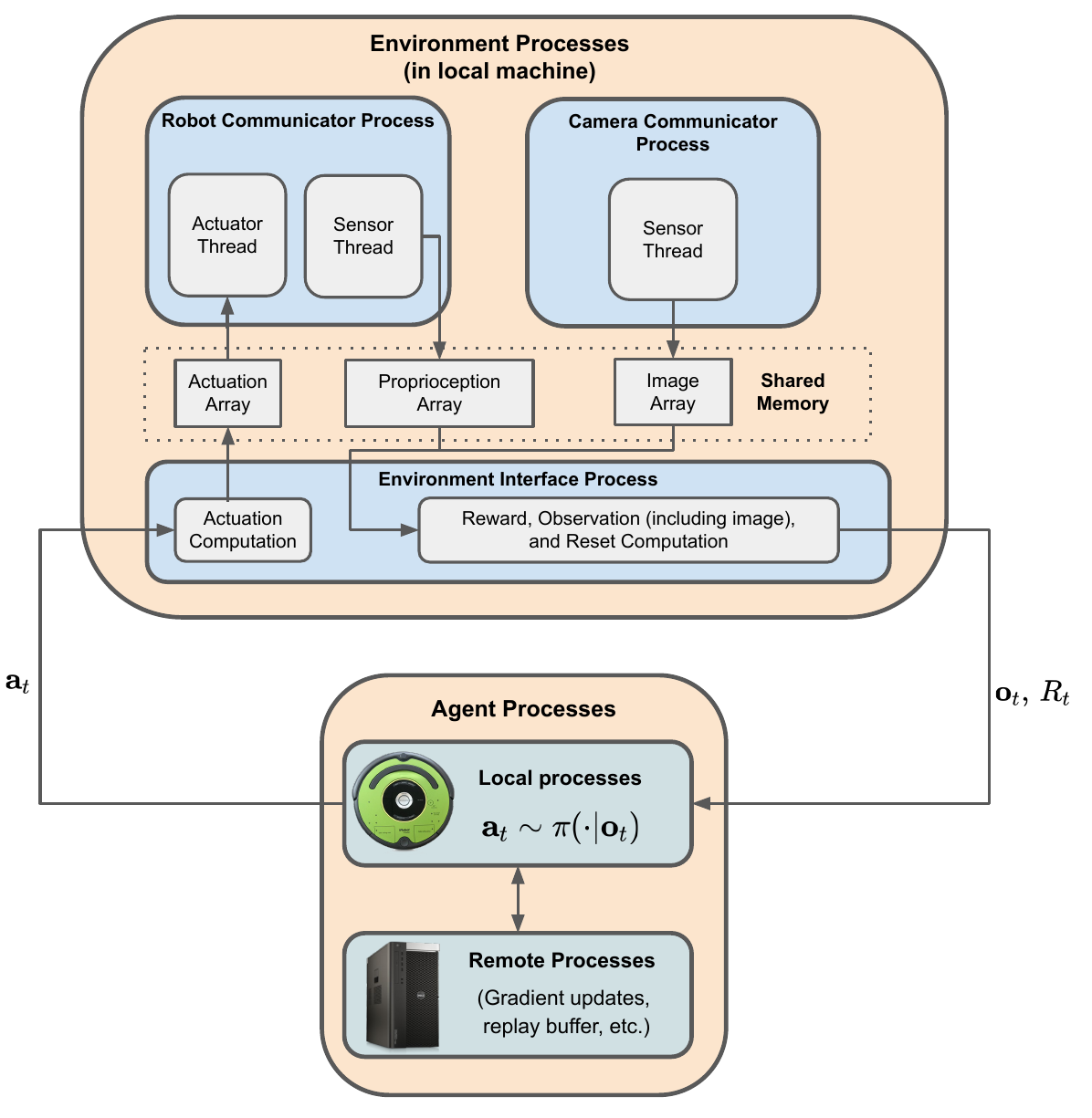}
\caption{Our proposed system ReLoD distributes computations of a learning system between a local and a remote computer.}
\label{fig:ROA_env_agent_interaction}
\end{figure}

Reinforcement Learning (RL) is a natural way of formulating real-time control learning tasks.
Although many deep RL methods have been developed to solve complex motor control problems
(Schulman et al.\ 2017, Abdolmaleki et al.\ 2018, Haarnoja et al.\ 2018)
they do not easily extend to the real-time learning setting that operates under time and resource constraints, for example, in quadrotors and mobile robot bases.
While approaches including learning from demonstrations  (Gupta et al.\ 2016, Vasan \& Pilarski 2017), sim-to-real (Peng et al.\ 2018, Bousmalis et al.\ 2018), and offline RL (Levine et al.\ 2020) have been used to develop pre-trained agents, there has been relatively little interest in studying real-time learning in the real world.

State-of-the-art RL algorithms such as SAC are computationally intensive, and hence for real-time robotic control, they go together with a computationally powerful computer tethered to the robot (Yuan \& Mahmood 2022).
On the other hand, a robot agent deployed in the real world typically uses a tethered resource-limited computer and a wirelessly connected workstation (Haarnoja et al.\ 2019, Bloesch et al.\ 2021).
In this paper, we use \emph{local} to refer to the computer tethered to the robot and \emph{remote} to refer to the wirelessly connected computer.
Computations of a learning system using these two computers can be distributed in different ways (e.g., see Fig.\ \ref{fig:ROA_env_agent_interaction}).
However, it is unclear how much the performance of a learning system is impacted due to the wireless connection or resource limitations.
Moreover, prior works do not systematically study distributions of computations between local and remote computers nor suggest how to achieve an effective distribution.

In this paper, we develop two vision-based tasks using a robotic arm and a mobile robot, and propose a real-time RL system called the \emph{Re}mote-\emph{Lo}cal \emph{D}istributed (ReLoD) system. 
Similarly to Yuan and Mahmood's (2022) work, ReLoD  parallelizes computations of RL algorithms to maintain small action-cycle times and reduce the computational overhead of real-time learning. But unlike the prior work, it is designed to utilize both a local and a remote computer.
ReLoD supports three different modes of distribution: \emph{Remote-Only} that allocates all computations on the remote computer,  \emph{Local-Only} that allocates all computations on the local computer, and \emph{Remote-Local} that carefully distributes the computations between the two computers in a specific way.
Our results show that the performance of SAC on a tethered resource-limited computer drops substantially compared to its performance on a tethered powerful workstation.
Surprisingly, when all computations of SAC are deployed on a wirelessly connected powerful workstation, the performance does not improve notably, which contradicts our intuition since this mode fully utilizes the workstation.
On the other hand, SAC's Remote-Local mode consistently improves its performance by a large margin on both tasks, which indicates that a careful distribution of computations is essential to utilize a powerful remote workstation.
However, the Remote-Local mode only benefits computationally expensive and sample efficient methods like SAC since the relatively simpler learning algorithm PPO learns similar policies in all three modes.
We also notice that the highest average return attained by PPO is about one-third of the highest average return attained by SAC, which indicates that SAC is more effective in complex robotic control tasks.
Our system in the Local-Only mode can achieve a performance that is on par with a system well-tuned for a single computer (Yuan \& Mahmood 2022), though the latter overall learns slightly faster.
This property makes our system suitable for conventional RL studies as well.



\thispagestyle{empty}
\section{RELATED WORK}

A system comparable to ours is SenseAct, which provides a computational framework for robotic learning experiments to be reproducible in different locations and under diverse conditions (Mahmood et al.\ 2018b).
Although SenseAct enables the systematic design of robotic tasks for RL, it does not address how to distribute computations of a real-time learning agent between two computers, and the original work does not contain vision-based tasks.
We use the guiding principles of SenseAct to design the vision tasks and systematically study the effectiveness of different distributions of computations of a learning agent. 

Krishnan et al.\ (2019) introduced an open-source simulator and a gym environment for quadrotors. 
Given that these aerial robots need to accomplish their tasks with limited onboard energy, it is prohibitive to run current computationally intensive RL methods on the onboard hardware. 
Since onboard computing is scarce and updating RL policies with existing methods is computationally intensive, they carefully designed policies considering the power and computational resources available onboard. 
However, they focused on sim-to-real techniques for learning, making their approach unsuited for real-time learning.

Nair et al.\ (2015) proposed a distributed learning architecture called the GORILA framework that mainly focuses on using multiple actors and learners to collect data in parallel and accelerate training in simulation using clusters of CPUs and GPUs. 
GORILA is conceptually akin to the DistBelief (Dean et al.\ 2012) architecture. 
In contrast to the GORILA framework, our system focuses primarily on how best to distribute the computations of a learning system between a resource-limited local computer and a powerful remote computer to enable effective real-time learning.
In addition, the GORILA framework is customized to Deep Q-Networks (DQN), while our system supports two different policy gradient algorithms using a common agent interface.

Lambert et al.\ (2019) used a model-based reinforcement learning approach for high-frequency control of a small quadcopter. 
Their proposed system is similar to our Remote-Only mode. 
A recent paper by Smith et al.\ (2022) demonstrated real-time learning of walking gait from scratch on a Unitree A1 quadrupedal robot on various terrains. Their real-time synchronous training of SAC on a laptop is similar to our Local-Only mode. 
The effectiveness of both these approaches on vision-based tasks is untested.

Bloesch et al.\ (2021) used a distributed version of Maximum aposteriori Policy Optimization (MPO) (Abdolmaleki et al.\ 2018) to learn a vision-based control policy that can walk with Robotis \href{https://emanual.robotis.com/docs/en/platform/op3/introduction/}{OP3 bipedal robots}. 
The robot's onboard computer samples actions and periodically synchronizes the policy's neural network weights with a remote learning process at the start of each episode. Haarnoja et al.\ (2019) also proposed a similar asynchronous learning system tailored to learn a stable gait using SAC and the minitaur robot (Kenneally et al.\ 2016). 
These tasks do not use images.
Although their proposed systems are similar to our Remote-Local mode, these two papers aim at solving tasks instead of systematically comparing different distributions of computations of a learning agent between a resource-limited computer and a powerful computer. 
In addition, their systems are tailored to specific tasks and algorithms and are not publicly available, while our system is open-source, task-agnostic, and compatible with multiple algorithms. 

\section{Background}

Reinforcement learning is a setting where an agent learns to control through trial-and-error interactions with its environment. 
The agent-environment interaction is modeled with a Markov Decision Process (MDP), where an agent interacts with its environment at discrete timesteps.
At the current timestep ${t}$, the agent is at state $S_t \in \mathcal{S}$, where it takes an action $A_t \in \mathcal{A}$ using a probability distribution $\pi$ called a \emph{policy}: $A_t\sim\pi(\cdot | S_t)$. 
At the subsequent timestep ${t+1}$, the agent receives the next state $S_{t+1}$ and reward $R_{t+1}$ according to a transition probability density function $S_{t+1}, R_{t+1} \sim p(\cdot, \cdot | S_t, A_t)$.
An episode ends when the agent arrives at a terminal state.
The state space $\mathcal{S}$ and the action space $\mathcal{A}$ of the tasks used in this paper are continuous. 

\subsection{Soft Actor-Critic}
Soft Actor-Critic (SAC) is an off-policy RL algorithm where the policy is trained to maximize a trade-off between the expected return and the policy's entropy (Haarnoja et al.\ 2018). 
SAC learns a policy $\pi_\phi(A_t|S_t)$ by maximizing the following objective:

\begin{align*}
    J_\pi (\phi) &= \mathbb{E}_{S_t \sim \mathcal{D}} \left[ \mathbb{E}_{A_t \sim \pi_\phi} [\alpha \log{\pi_\phi (A_{t} | S_{t})} - Q_\theta (S_{t}, A_{t})] \right],
\end{align*}
which is based on an entropy-augmented version of the standard RL objective (Ziebart 2010).
Here $\phi$ denotes the policy parameters, and $Q_\theta (S_{t}, A_{t})$, parameterized by $\theta$, denotes an estimate of the Soft-Q function $q_\pi$ defined as $q_\pi(s, a) = E_\pi\left[\sum_{t=0}^{T-1} R_{t+1} - \alpha\sum_{t=1}^{T-1}\log\pi(A_t|S_t)\big|S_0=s, A_0=a\right]$.

The parameters of the soft Q-function are trained to minimize the following objective:
\begin{align*}
    J_Q (\theta) &= \mathbb{E}_D \left[ \frac{1}{2} \left( R_{t+1} + \gamma V_{\Tilde{\theta}}(S_{t+1})-Q_\theta (S_t, A_t)  \right)^2 \right],
\end{align*}
where $V_{\Tilde{\theta}}$ is defined as
\begin{align*}
    V_{\Tilde{\theta}}(S_{t}) &= \mathbb{E}_{\pi} \left[ Q_\theta (S_{t}, A_{t}) - \alpha \log{\pi_\phi (A_{t} | S_{t})} \right],
\end{align*}
with $\Tilde{\theta}$ being the parameters of the target network.
In this paper, we use the undiscounted episodic formulation. 
In other words, $\gamma$ is set to 1.

\subsection{Proximal Policy Optimization (PPO)}
PPO (Schulman et al.\ 2017) is an on-policy policy gradient algorithm that alternates between collecting data via interactions with the environment and optimizing the following clipped surrogate objective:
\begin{align*}
 L_{\theta}^\text{CLIP} = &\mathbb{E}_{S_t,A_t \sim \pi_{\theta_{old}}} [ \min (\rho_\theta(A_t|S_t) {H}_{\theta_\text{old}}(S_t, A_t), \\ 
 & \quad\text{clip}(\rho_\theta(A_t|S_t), 1 - \epsilon, 1 + \epsilon) {H}_{\theta_\text{old}}(S_t, A_t))],
\end{align*}
where $\rho_\theta(A_t|S_t) = \frac{\pi_\theta(A_t \vert S_t)}{\pi_{\theta_\text{old}}(A_t \vert S_t)}$ is the action probability ratio between the new policy and the old policy, and $H_{\theta_\text{old}}(S_t, A_t)$ is an estimate of the advantage function. 
The function $\text{clip}(\rho_\theta(A_t|S_t), 1 - \epsilon, 1 + \epsilon)$ clips the probability ratio to be no more than $1 + \epsilon$ if the advantage is positive and no less than $1 - \epsilon$ if the advantage is negative. 
The policy is optimized by running several iterations of stochastic gradient ascent at each update. 

\thispagestyle{empty}
\section{TASKS}
\label{section:tasks}
For this paper, we develop two visuomotor robotic tasks, named \emph{UR5-VisualReacher} and \emph{Create-Reacher}, that are not easy to solve using simple RL approaches since they require tens of thousands of images to learn good state representations and control policies.
The robots we used in this paper are the \textbf{UR5} industrial robot arm and iRobot \textbf{Create2} mobile robot. 
All environments are designed using the guidelines proposed by Mahmood et al.\ (2018a) to reduce the computational overhead and take advantage of the multiprocessing capability of modern computers for handling different transmission rates.
\subsection{UR5-VisualReacher}
This environment is the same task proposed by Yuan and Mahmood (2022). 
It is an image-based task that aims to move a UR5 arm's fingertip to a random target designated by a red (the target color) blob on a monitor. 
The movement of the UR5 arm is restricted within a bounding box to prevent collisions and accidental damage.
The action space is the desired angular velocities for the five joints between $[-0.7, 0.7]$rad/s. 
The observation vector includes joint angles, joint angular velocities, three consecutive images of dimension $160\!\times\!90\!\times\!3$ taken by a camera, and the previous action.
Every 40ms, the SenseAct environment process actuates the five joints and receives the next observation by sending low-level control commands. 
The SenseAct communicator process transmits joint data every 8ms and image data every $33.\Bar{3}$ms to the environment process.
Since this environment is inherently non-Markovian, we stack three images to provide the partial history to the learning agent. 
Each episode lasts 4 seconds (i.e., 100 timesteps). The reward function is designed to encourage the following behaviors: 1) moving the fingertip closer to the target while keeping the target centralized, and 2) avoiding abrupt twisting of the joints. 
Specifically, the reward function is defined as follows (Yuan \& Mahmood (2022)):
\[
    r_t=\alpha \frac{1}{hw}M \circ W-\beta \left(\bigg|\pi-\sum_{n=1}^{3}\omega_n\bigg|+\bigg|\sum_{n=4}^{5}\omega_n\bigg|\right),
\]
where $h$ is the height and $w$ is the width of the image observation, $M$ is a mask matrix whose element $M_{i,j}$ is 1 if the target color is detected at location $(i,j)$ and 0 otherwise, $W$ is a constant weight matrix whose elements range from 0 to 1 to encourage centralization of the target, and $\omega$ is the vector of the five joints' angles. 
The coefficient $\alpha$ and $\beta$ weigh the importance of the two behaviors. 
In this paper, we set $\alpha=800$ and $\beta=1$. During a reset, a new random target will be displayed on the monitor for the next episode, and the UR5 arm will be set to a predefined posture. 
The environment runs on a Lenovo Ideapad Y700 laptop.

\begin{figure*}
\centering
    \begin{subfigure}{.95\columnwidth}
        \includegraphics[width=\columnwidth]{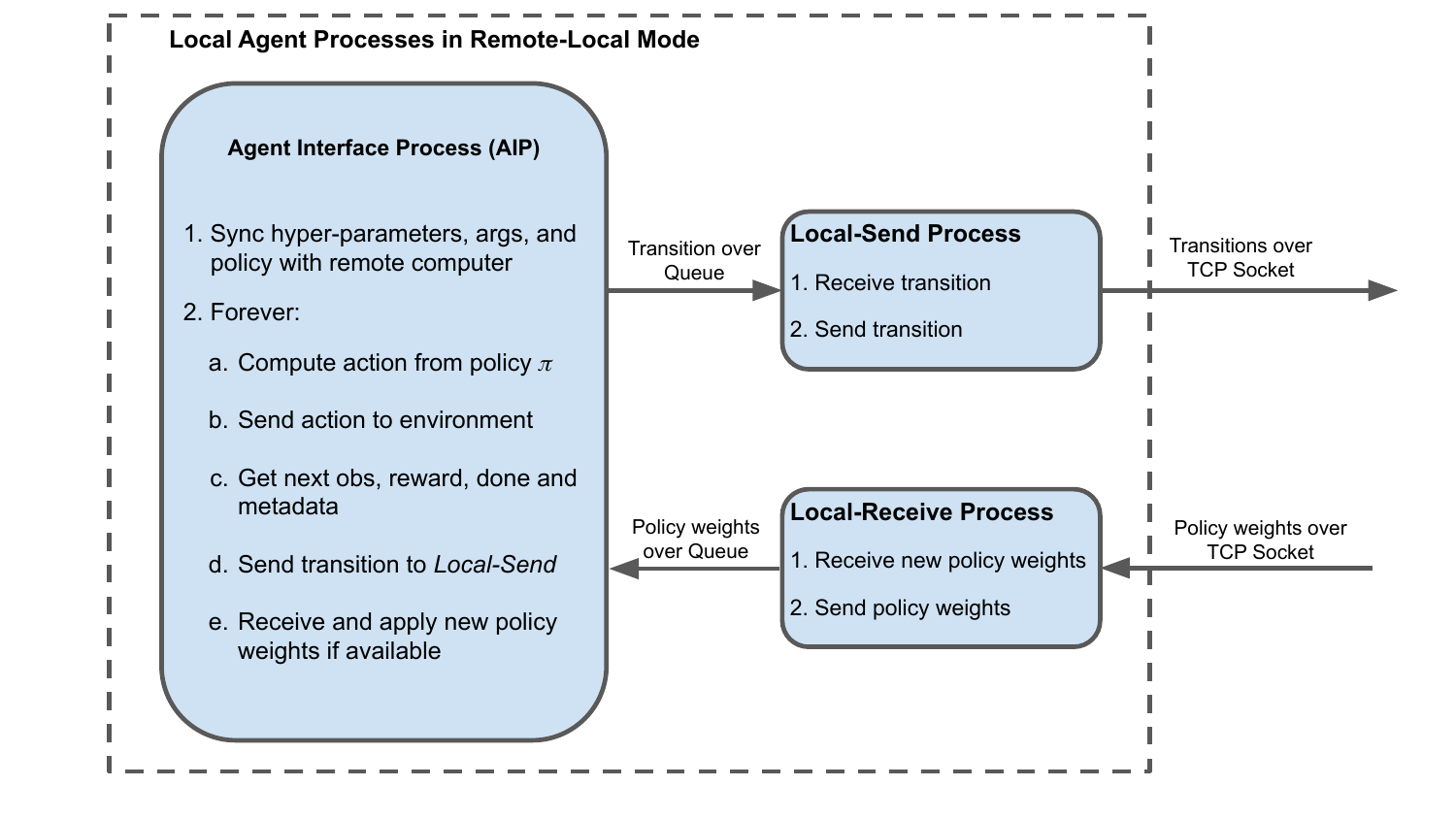}
        \caption{}
        \label{fig:local_agent}
    \end{subfigure}\hfill%
    \begin{subfigure}{.95\columnwidth}
        \includegraphics[width=\columnwidth]{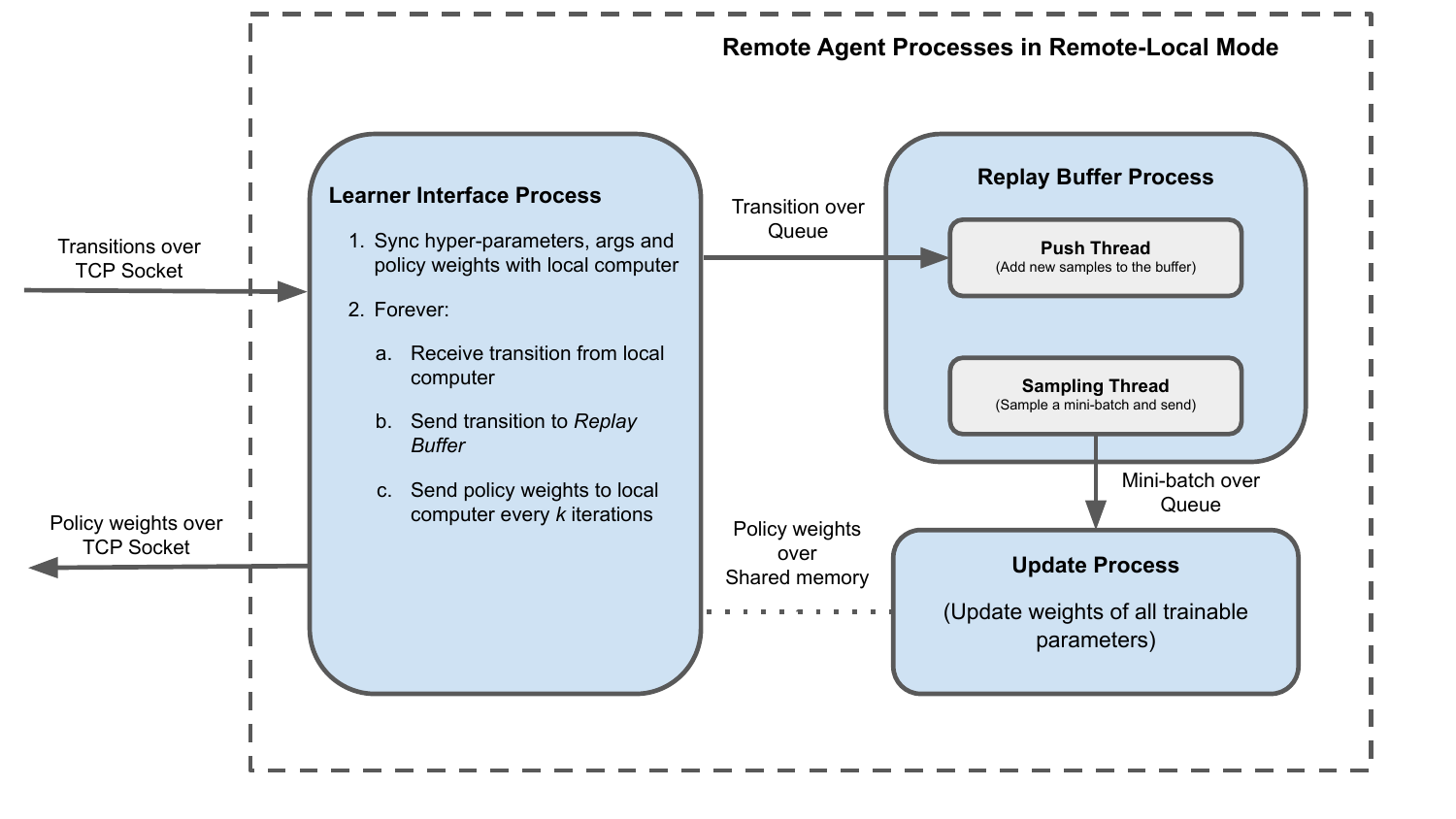}
        \caption{}
        \label{fig:remote_agent}
    \end{subfigure}
\caption{The SAC-Agent in the Remote-Local Mode. 
The lengths of the transition queues on the local computer and the remote computer are set to the maximum episode length to avoid losing data.}
\label{fig:remote_local}
\end{figure*}

\begin{table*}
  \begin{center}
    \begin{tabular}{|c|c|c|c|c|} 
        \hline
          &  \textbf{Remote-Local} & \textbf{Remote-Only} & \textbf{Local-Only} & \textbf{Period of Iteration}\\ \hline
        \textbf{Agent Interface Process (AIP)} & Local & Local & Local & action cycle time \\ \hline 
        \textbf{Action computation}  & Included in AIP & Included in LIP  & Included in AIP & action cycle time \\ \hline 
        \textbf{Local-Send Process} & Local & N/A  & N/A & action cycle time\\ \hline 
        \textbf{Local-Receive Process} & Local & N/A  & N/A & Every $k$ iterations of LIP's loop\\ \hline 
        \textbf{Learner Interface Process (LIP)} & Remote & Remote & Included in AIP & action cycle time\\ \hline 
        \textbf{Replay Buffer Process} & Remote & Remote & Local & hardware dependent \\ \hline 
        \textbf{Update Process} & Remote & Remote & Local & hardware dependent \\ \hline
    \end{tabular}
  \end{center}

  \caption{A list of all processes and their distributions between the local and the remote computers in various operational modes of the ReLoD system.}
  \label{tab:1}
\end{table*}

\subsection{Create-Reacher}
This image-based task aims to move Create2 as soon as possible to a target designated by a green piece of paper attached to a wall. 
The target is thought to be reached if it occupies more than 12\% pixels of the current image. 
The size of the image is $160\!\times\!120\!\times\!3$. 
The movement of Create2 is limited by an arena of size $110 \text{cm} \times 70 \text{cm}$ built by wood and cardboard. 
The action is the desired speeds for the two wheels, limited to $[-150, 150]$mm/s to prevent damage to Create2. 
The observation vector includes the values of the six wall sensors, three consecutive images taken by a camera, and the previous action. 
Every 45ms, the SenseAct environment process actuates the two wheels and receives the next observation by sending low-level control commands. 
The SenseAct communicator process transmits the sensory data every 15ms and the image data every $33.\Bar{3}$ms to the environment process. 
Similar to UR5-VisualReacher, we stack three images to give the partial history to the learning agent. 
Each episode has a maximum length of 30 seconds. 
The current episode will terminate earlier if Create2 reaches the target. The reward is $-1$ per step to encourage shorter episodes.
The reset routine will bring Create2 to a random location in the arena. 
During a reset, Create2 first moves backward and reorients itself in a random direction. 
Our reset routine also handles recharging when the battery life falls below a specified threshold. 
This environment runs on a Jetson Nano computer.

\thispagestyle{empty}
\section{
The proposed ReLoD system
}
\label{section: the system}

In this section, we describe our proposed ReLoD system, which distributes the computational components of a real-time RL agent between a local computer and a remote computer.
Most deep RL systems have three computationally expensive components: 1) Action Computation, 2) Policy Updating, and 3) Minibatch Sampling. 
ReLoD spawns parallel processes for each of the three components and runs them in an asynchronous, distributed manner to better utilize the computational resources of the local and remote computers.
It also creates additional inter-process communication sockets to enable this distributed and parallel architecture.
ReLoD supports three different modes of operation: \textit{Remote-Only}, \textit{Remote-Local}, and \textit{Local-Only.}
The mode of operation determines how the various processes are distributed between the local computer and the remote computer.
ReLoD is designed to work with various RL algorithms.
Users can easily add more RL algorithms to the system by implementing the agent interface. 
In this paper, we implement two state-of-the-art RL algorithms: SAC and PPO. 
We call the resulting agents \emph{SAC-Agent} and \emph{PPO-Agent}, respectively. 

Fig.\ \ref{fig:remote_local} outlines the structure of the SAC-Agent in the Remote-Local mode.
There are three processes, called \textit{Agent-Environment Interface}, \textit{Local-Send}, and \textit{Local-Receive}, that run in parallel on the local computer. 
The Agent-Environment Interface process computes the actions to take in the environment (except in Remote-Only mode given in Table  \ref{tab:1}).
Three more processes, called \textit{Learner Interface}, \textit{Replay Buffer}, and \textit{Update},  run in parallel on the remote computer. 
The Replay Buffer process samples mini-batches, and the Update process updates policies.
By distributing the three most computationally expensive components of SAC on two different computers, ReLoD takes advantage of both computers.
On the other hand, in the Remote-Only mode, all agent computations happen on the remote computer, including action computation, and the local computer relays observations and actions between the remote computer and the robot.
More details about the process distribution for all three modes of the SAC-Agent are given in Table \ref{tab:1}.

Note that the learning algorithm determines what processes to use in each mode.
For example, the Replay Buffer and the Update processes are specific to our asynchronous SAC implementation adapted from Yuan and Mahmood (2022). 
The PPO-Agent does not have them due to the nature of the algorithm. It updates policies in the Learner Interface process instead.

\thispagestyle{empty}
\section{EXPERIMENTS \& RESULTS}
\label{section:expts}

\begin{figure*}
\centering
    \begin{subfigure}{.32\textwidth}
        \includegraphics[width=\columnwidth]{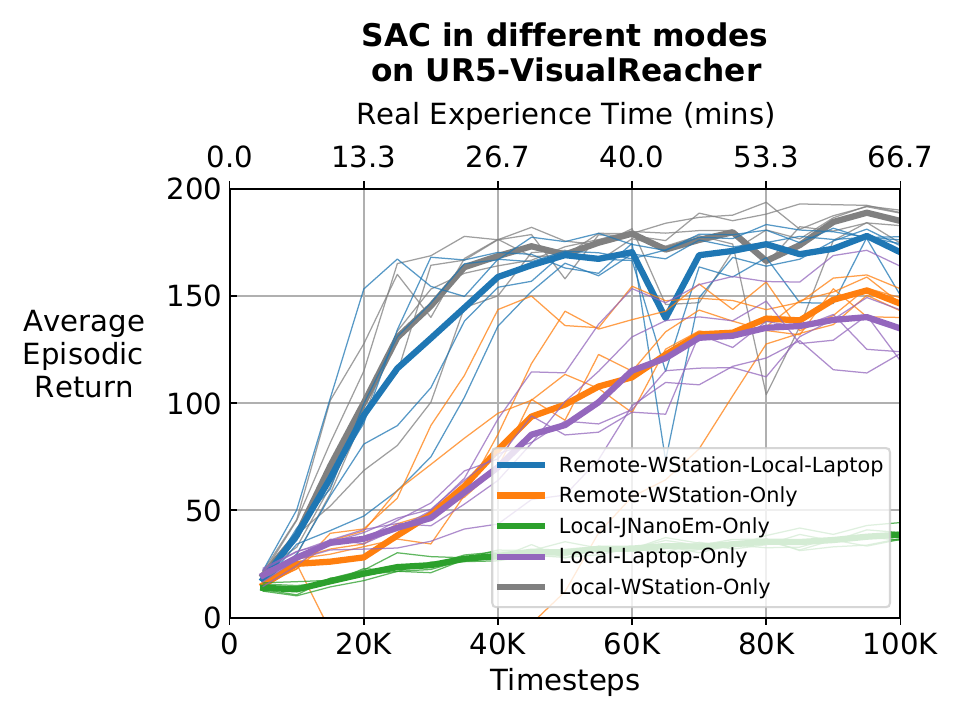}
        \caption{}
        \label{fig:ur5_sac}
    \end{subfigure}
    \begin{subfigure}{.32\textwidth}
        \includegraphics[width=\columnwidth]{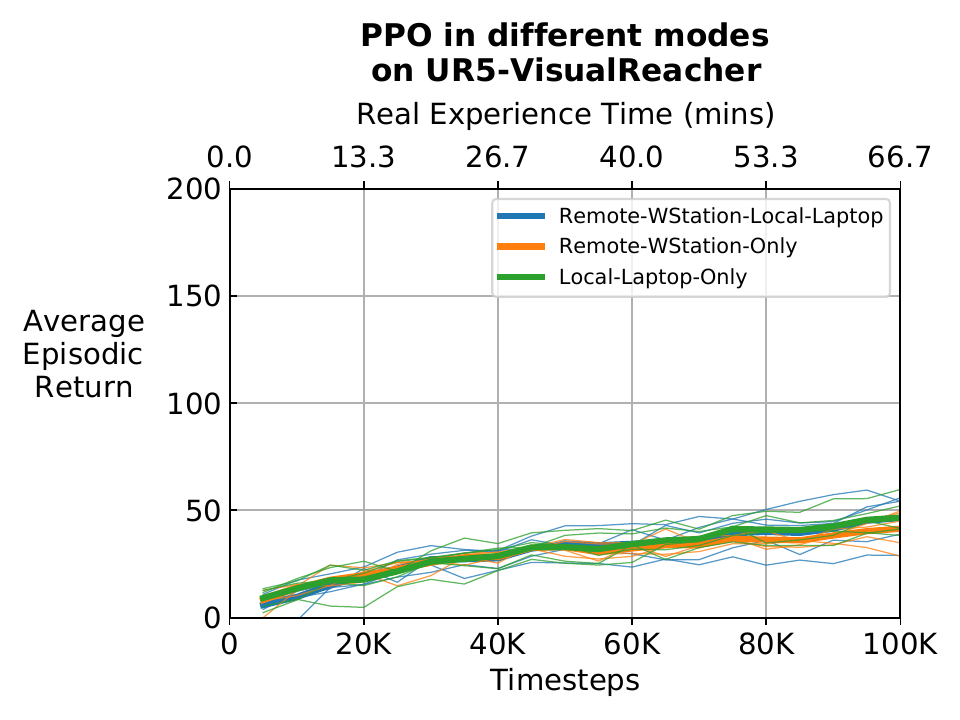}
        \caption{}
        \label{fig:ur_ppo}
    \end{subfigure}
        \begin{subfigure}{.32\textwidth}
        \includegraphics[width=\columnwidth]{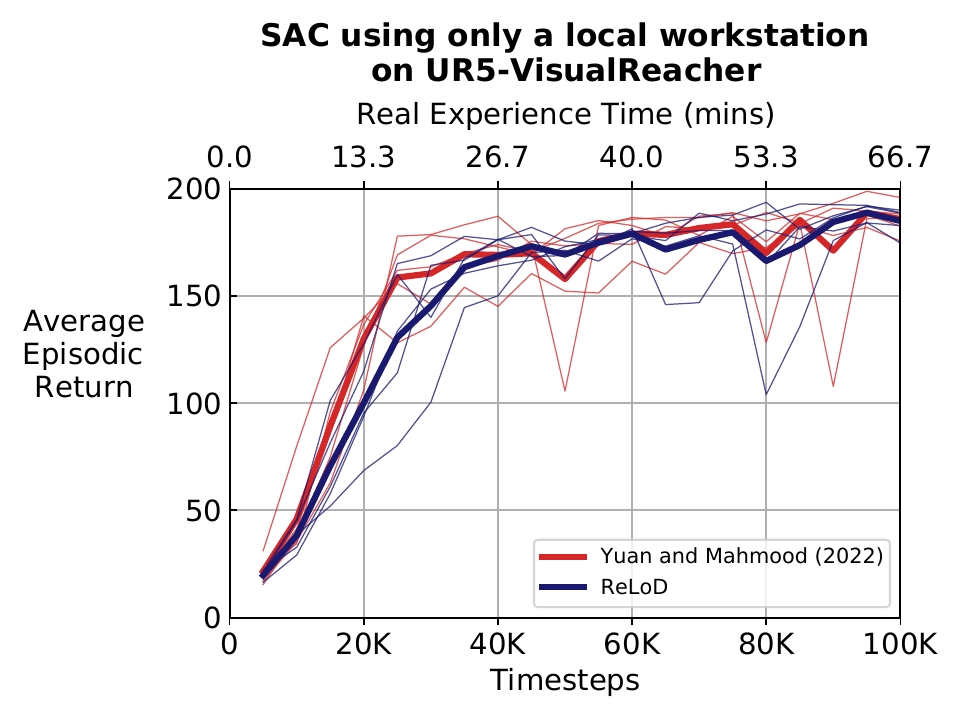}
        \caption{}
        \label{fig:yufeng_ours}
    \end{subfigure}
    \caption{Comparison of the learning performance across three operational modes of ReLoD on UR5-VisualReacher. 
    The wide lines for each mode are averaged over five independent runs.
    The real experience time in our plots does not include any additional components required to run an experiment, such as resetting the environment or recharging the robot. 
    Fig.\ \ref{fig:ur5_sac} shows the comparison of the learning performance of the SAC-Agent. 
    Fig.\ \ref{fig:ur_ppo} shows the comparison of the learning performance of the PPO-Agent. 
    Fig.\ \ref{fig:yufeng_ours} shows the comparison of the Local-Only mode of the SAC-Agent with an off-the-shelf asynchronous SAC implementation proposed by \protect Yuan and Mahmood (2022) on a powerful workstation.}
    \label{fig:ur5_curves}
\end{figure*}
In this paper, we first compared the performance of Local-Only SAC between a resource-limited local computer and a powerful local workstation, referred to as \textit{Local-WStation-Only}, to evaluate the impact on performance due to resource limitations.
The default hyper-parameters provided by Yuan and Mahmood (2022) were used in this paper.
We used the IdeaPad Y700 laptop as the resource-limited computer.
It has an Intel i7-6700HQ CPU, an NVidia GTX 960M GPU, and 16GB memory.
We reduced SAC's mini-batch size to 64 on the laptop to fit the hardware constraints. 
The maximum allowable buffer size was also reduced compared to the workstation, within which size of 16000 performed well.

Considering that heavily resource-limited computers like Jetson Nano are commonly used as the local controlling computers for small robots, we also aimed to investigate how SAC would have performed on Jetson Nano.
However, when ReLoD was configured to Local-Only SAC, Jetson Nano could not run the system fully onboard as its hardware and software architectures were not optimized for updating network weights.
Jetson Nano took about half a second to complete one network update on average and became unresponsive after a couple of updates.
To overcome the architectural limitations and get an empirical upper bound of performance on such a heavily resource-limited computer, we emulated the capability of Jetson Nano with the laptop by restricting its update rate and available memory.
Thus, two Local-Only variants of SAC were tested on the laptop: 
\begin{itemize}
    \item \emph{Local-Laptop-Only}: This variant fully utilizes the computational resources of the laptop. 
    The policy is updated at the fastest allowable speed, that is, back-to-back.
    \item \emph{Local-JNanoEm-Only}: This variant emulates the capability of Jetson Nano. 
    Since the average time required for one SAC update on Jetson Nano is $500ms$ and the action cycle time of UR5-VisualReacher is $40ms$, the policy is updated once per 12 steps on the laptop to best match the capability of Jetson Nano. The buffer size is set to 16000, a maximum that fits Nano's 4GB RAM.
\end{itemize}
We used UR5-VisualReacher for this comparison as mounting the laptop to Create2 was impractical.
The workstation for UR5-VisualReacher has an AMD Ryzen Threadripper 2950 processor, an NVidia 2080Ti GPU, and 128G memory.
The results are shown in Fig \ref{fig:ur5_sac}.

Fig \ref{fig:ur5_sac} shows that the performance of SAC on the resource-limited laptop dropped by about 28\% on average compared to that of SAC on the workstation.
Moreover, the Local-JNanoEm-Only variant struggled to learn a comparable policy within the same time frame.
Note that SAC's Local-WStation-Only configuration attained the highest return among all our results.

Since tethering a workstation to a mobile robot is inconvenient in most cases, we investigated how to utilize the resource of a remote workstation to compensate for the performance loss due to the resource-limited local computer. 
We used the most common setting in practice in which the robot is tethered to a small resource-limited computer and wirelessly connected to a powerful workstation.
To get better coverage of RL algorithms, two typical algorithms, SAC and PPO, were tested.
PPO was adapted from Achiam (2018), and we extended it to support learning with images.
The default hyper-parameters provided by Achiam (2018) were used for PPO.

SAC is resource-demanding since it updates the policy every step and requires a large replay buffer to store all its past experiences. PPO is relatively simpler since it only updates the policy every few thousand steps, and its buffer needs to hold only a few episodes generated on-policy.
The SAC-Agent was tested on UR5-VisualReacher and Create-Reacher, while the PPO-Agent was only tested on UR5-VisualReacher.
As the local computer, we used a Jetson Nano 4GB for Create-Reacher and the laptop for UR5-VisualReacher.
The workstation for Create-Reacher has an AMD Ryzen Threadripper 3970 CPU, an NVidia 3090 GPU, and 128G memory.

\begin{figure}
    \centering
    \begin{subfigure}{0.33\textwidth}
        \includegraphics[width=\columnwidth]{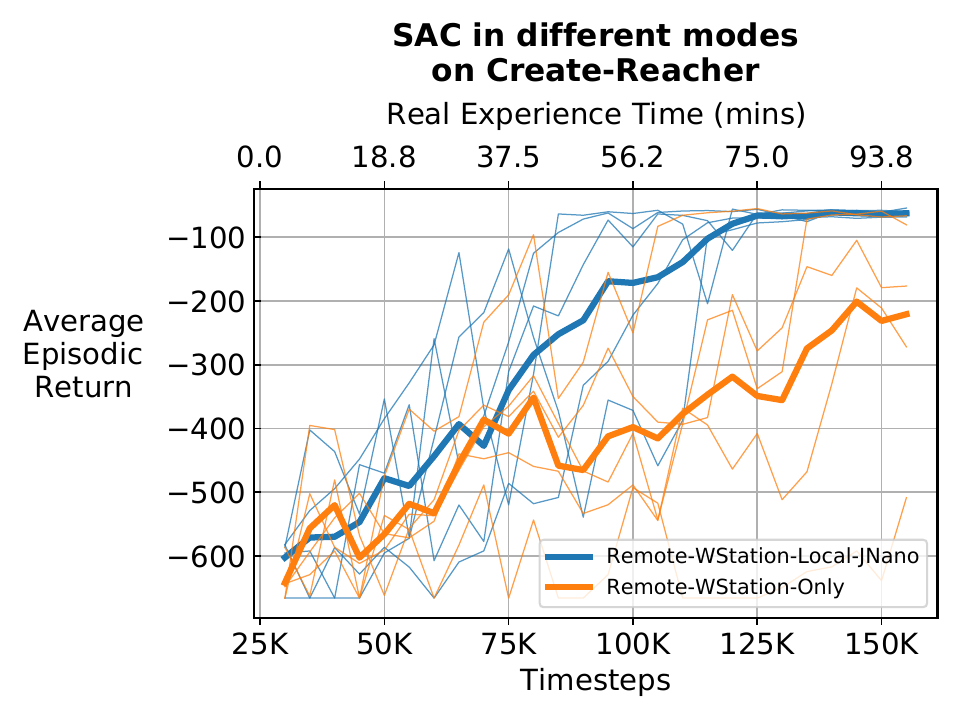}
    \end{subfigure}
    \caption{Comparison of learning performance across two modes of SAC-Agent on Create-Reacher. 
    The wide lines for each mode are averaged over five independent runs.}
    \label{fig:create_sac}
\end{figure}

The learning curves of the three modes of the SAC-Agent on UR5-VisualReacher and Create-Reacher are shown in Fig.\ \ref{fig:ur5_sac} and Fig.\ \ref{fig:create_sac}, respectively.
Note that for the same reason mentioned above, SAC-Agent's Local-Only mode was not run for Create-Reacher.
Fig \ref{fig:ur5_sac} reveals a counterintuitive result that SAC's Remote-Only mode, referred to as \emph{Remote-WStation-Only} in the Figures, barely improved its performance over the Local-Only mode, which used a resource-limited laptop.
Meanwhile, the Remote-Only mode also exhibited higher variance in the overall learning performance on both tasks.
This is surprising since this mode fully utilizes all resources of the remote workstation.
We attribute the relatively poor performance of the Remote-Only mode to the variable latency in communicating actions over WiFi, which will not occur in the other two modes, as the Remote-Only mode computes actions on a remote computer.
Although the Remote-Local mode also encounters latency due to WiFi communication, the effect is much less severe as it only involves the transfer of policies and buffer samples, while action computation still happens locally.
Our results are in line with Mahmood et al.'s (2018a)
results 
showing that delays occurring closer to robot actuations can be substantially more detrimental to performance compared to delays occurring further from actuations.
A real-time learning system can get affected by delays at least through two pathways: inference for action computation and learning update.
When a delay occurs within the system, it may spare the more important inference pathway, as in our case. 
Whereas when a delay occurs closer to the robot actuation, that is, at the periphery of the learning system, it affects both pathways.

Fig.\ \ref{fig:ur5_sac} and Fig.\ \ref{fig:create_sac} suggest that SAC-Agent's Remote-Local mode, referred to as \emph{Remote-WStation-Local-Laptop} and \emph{Remote-WStation-Local-JNano} in the Figures, consistently compensated for the performance loss on both tasks.
As Fig.\ \ref{fig:ur5_sac} indicates, the highest average return attained by SAC-Agent's Remote-Local mode is more than 90\% of the highest average return attained by SAC in the Local-WStation-Only configuration, whose performance is the best.
However, the Remote-Local mode only benefits computationally intensive algorithms like SAC, as Fig.\ \ref{fig:ur_ppo} shows that PPO's performance is nearly the same across the three modes.

By comparing Fig.\ \ref{fig:ur5_sac} and Fig.\ \ref{fig:ur_ppo}, we notice that SAC significantly outperformed PPO on complex robotic tasks.
All distribution modes of the SAC-Agent except the Local-JNanoEm-Only attained more than twice the return attained by PPO in a shorter time frame.
Nevertheless, when all computations of SAC happen on a computer incapable of performing frequent policy updates, indicated by Local-JNanoEm-Only, SAC's performance degrades substantially to a degree comparable to that of PPO on the same computer.

Finally, on a powerful tethered workstation, we compare the performance of Local-Only SAC to that of the system proposed by Yuan and Mahmood (2022), which is well-tuned for a single computer. 
This test is to determine if the performance of the Local-Only mode of our system is comparable to the off-the-shelf RL algorithm implementation given adequate resources. 
The environment was UR5-VisualReacher, and five independent runs were performed for each system.
The comparison results are shown in Fig.\ \ref{fig:yufeng_ours}. 
Both methods achieved similar overall learning performance. 
Thus, our proposed system can leverage the power of deep RL algorithms effectively, and as a result, it is also suitable for conventional RL research.

\thispagestyle{empty}
\section{Conclusion}
In this paper, we introduced the ReLoD system for learning to control robots by a real-time RL agent that distributes its computations between a local and a remote computer.
In our experiments with ReLoD, we demonstrated that when SAC ran on a resource-limited computer, its performance could be dramatically reduced. 
Moreover, solely deploying all computations of SAC on a wirelessly connected remote workstation may not improve performance due to latency.
On the other hand, our results suggest that performing local action computation and remote policy updating compensates substantially for SAC's performance loss.
However, this distribution may not benefit all RL algorithms since PPO's performance was nearly unaffected by the distribution mode.

ReLoD provides a useful tool and guidance for real-time RL in vision-based robotic control.
Computationally intensive algorithms like SAC can best utilize ReLoD's Remote-Local mode. 
Moreover, ReLoD's Local-Only mode is suitable for conventional RL research as its performance was shown to be on par with a well-tuned single-computer system.
However, due to the latency in communicating actions, ReLoD's Remote-Only mode should only be used when the local computer cannot compute actions within the action cycle time.

\section{Acknowledgements}
We gratefully acknowledge funding from the CCAI Chairs program, the RLAI laboratory, Amii, and NSERC of Canada. We also thankfully acknowledge the donation of the UR5 arm from the Ocado Group.

\addtolength{\textheight}{-4cm}   





\thispagestyle{empty}
\section{References}
\hangindent=2em
\hangafter=1
\noindent Abdolmaleki, A., Springenberg, J. T., Tassa, Y., Munos, R., Heess, N., \& Riedmiller, M. (2018). Maximum a posteriori policy optimisation. In \textit{International Conference
on Learning Representations.}

\hangindent=2em
\hangafter=1
\noindent Achiam, J. (2018). Spinning Up in Deep Reinforcement Learning.

\hangindent=2em
\hangafter=1
\noindent Bloesch, M., Humplik, J., Patraucean, V., Hafner, R., Haarnoja, T., Byravan, A., . . . others (2021). Towards real robot learning in the wild: A case study in bipedal locomotion. In \textit{5th Annual Conference on Robot Learning.}

\hangindent=2em
\hangafter=1
\noindent Bousmalis, K., Irpan, A., Wohlhart, P., Bai, Y., Kelcey, M., Kalakrishnan, M., . . . others (2018). Using simulation and domain adaptation to improve efficiency of deep
robotic grasping. In \textit{2018 IEEE international conference on robotics and automation (ICRA) (pp. 4243–4250).}

\hangindent=2em
\hangafter=1
\noindent Dean, J., Corrado, G., Monga, R., Chen, K., Devin, M., Mao, M., ... \& Ng, A. (2012). Large scale distributed deep networks. \textit{Advances in neural information processing systems}, 25.

\hangindent=2em
\hangafter=1
\noindent Gupta, A., Eppner, C., Levine, S., \& Abbeel, P. (2016). Learning dexterous manipulation for a soft robotic hand from human demonstrations. In \textit{2016 IEEE/RSJ International Conference on Intelligent Robots and Systems (IROS)} (pp. 3786–3793).

\hangindent=2em
\hangafter=1
\noindent Haarnoja, T., Ha, S., Zhou, A., Tan, J., Tucker, G., \& Levine, S. (2019). Learning to walk via deep reinforcement learning. In \textit{Robotics: Science and Systems.}

\hangindent=2em
\hangafter=1
\noindent Haarnoja, T., Zhou, A., Abbeel, P., \& Levine, S. (2018). Soft actor-critic: Off-policy maximum entropy deep reinforcement learning with a stochastic actor. In \textit{International conference on machine learning} (pp. 1861–1870).

\hangindent=2em
\hangafter=1
\noindent Kenneally, G., De, A., \& Koditschek, D. E. (2016). Design principles for a family of direct-drive legged robots. \textit{IEEE Robotics and Automation Letters}, 1(2), 900–
907.

\hangindent=2em
\hangafter=1
\noindent Krishnan, S., Boroujerdian, B., Fu, W., Faust, A., \& Reddi, V. J. (2019). Air learning: An ai research platform for algorithm-hardware benchmarking of autonomous aerial robots. \textit{arXiv preprint arXiv:1906.00421.}

\hangindent=2em
\hangafter=1
\noindent Lambert, N. O., Drew, D. S., Yaconelli, J., Levine, S., Calandra, R., \& Pister, K. S. (2019). Low-level control of a quadrotor with deep model-based reinforcement
learning. \textit{IEEE Robotics and Automation Letters}, 4(4), 4224–4230.

\hangindent=2em
\hangafter=1
\noindent Laskin, M., Lee, K., Stooke, A., Pinto, L., Abbeel, P., \& Srinivas, A. (2020). Reinforcement learning with augmented data. \textit{Advances in Neural Information Processing Systems}, 33, 19884–19895.

\hangindent=2em
\hangafter=1
\noindent Levine, S., Kumar, A., Tucker, G., \& Fu, J. (2020). Offline reinforcement learning: Tutorial, review, and perspectives on open problems. \textit{arXiv preprint
arXiv:2005.01643.}

\hangindent=2em
\hangafter=1
\noindent Mahmood, A. R., Korenkevych, D., Komer, B. J., \& Bergstra, J. (2018a). Setting up a reinforcement learning task with a real-world robot. In \textit{2018 IEEE/RSJ International Conference on Intelligent Robots and Systems (IROS)} (pp. 4635-4640). IEEE.

\hangindent=2em
\hangafter=1
\noindent Mahmood, A. R., Korenkevych, D., Vasan, G., Ma, W., \& Bergstra, J. (2018b). Benchmarking reinforcement learning algorithms on real-world robots. \textit{In Confer-
ence on robot learning} (pp. 561–591).

\hangindent=2em
\hangafter=1
\noindent Nair, A., Srinivasan, P., Blackwell, S., Alcicek, C., Fearon,
R., De Maria, A., . . . others (2015). Massively parallel methods for deep reinforcement learning. \textit{arXiv preprint arXiv:1507.04296.}

\hangindent=2em
\hangafter=1
\noindent Peng, X. B., Andrychowicz, M., Zaremba, W., \& Abbeel, P. (2018). Sim-to-real transfer of robotic control with dynamics randomization. In \textit{2018 IEEE international conference on robotics and automation (ICRA)} (pp. 3803–3810).

\hangindent=2em
\hangafter=1
\noindent Ramstedt, S., \& Pal, C. (2019). Real-time reinforcement learning. \textit{Advances in neural information processing systems}, 32.

\hangindent=2em
\hangafter=1
\noindent Schulman, J., Wolski, F., Dhariwal, P., Radford, A., \& Klimov, O. (2017). Proximal policy optimization algorithms. \textit{arXiv preprint arXiv:1707.06347.}

\hangindent=2em
\hangafter=1
\noindent Smith, L., Kostrikov, I., \& Levine, S. (2022). A walk in the park: Learning to walk in 20 minutes with model-free reinforcement learning. \textit{arXiv preprint arXiv:2208.07860.}

\hangindent=2em
\hangafter=1
\noindent Vasan, G., \& Pilarski, P. M. (2017). Learning from demonstration: Teaching a myoelectric prosthesis with an intact limb via reinforcement learning. In \textit{2017 International Conference on Rehabilitation Robotics (ICORR)} (pp. 1457–1464).

\hangindent=2em
\hangafter=1
\noindent Yuan, Y., \& Mahmood, A. R. (2022). Asynchronous reinforcement learning for real-time control of physical robots. \textit{In 2022 IEEE international conference on
robotics and automation (ICRA).}

\hangindent=2em
\hangafter=1
\noindent Ziebart, B. D. (2010). Modeling purposeful adaptive behavior with the principle of maximum causal entropy. Carnegie Mellon University.

\newcommand{\hangin}{\goodbreak\hangindent=.3cm \noindent}

\newpage
\end{document}